\documentclass[10pt,twocolumn,letterpaper]{article}
\usepackage[T1]{fontenc}
\usepackage{iccv}              
\usepackage[pagebackref,breaklinks,colorlinks,allcolors=iccvblue]{hyperref}
\usepackage{graphicx}
\usepackage{multirow}
\usepackage{booktabs}
\usepackage{array}
\usepackage{booktabs}
\newcommand{\shadow}[1]{}
\def\s{\shadow}
\newcommand{\blue}[1]{\textcolor{black}{#1}}
\def\b{\blue}
\newcommand{\red}[1]{\textcolor{black}{#1}}
\def\r{\red}
\newcommand{\grn}[1]{\textcolor{black}{#1}}
\newcommand{\brown}[1]{\textcolor{black}{#1}}

\renewcommand{\red}[1]{{\color{red}#1}}

\definecolor{iccvblue}{rgb}{0.21,0.49,0.74}
\def\paperID{10955} 
\def\confName{ICCV}
\def\confYear{2025}

\title{Language-Guided Reinforcement Learning for Hard Attention in Few-Shot Learning}

\author{Bahareh Nikpour$^{1,2}$, Narges Armanfard$^{1,2}$\\
$^1$Department of Electrical and Computer Engineering, McGill University \\
$^2$Mila - Quebec AI Institute, Montreal, QC, Canada \\
{\tt\small bahareh.nikpour@mail.mcgill.ca, narges.armanfard@mcgill.ca}
}
\begin{document}
\maketitle
\begin{abstract}
\brown{Attention mechanisms have demonstrated significant potential in enhancing learning models by identifying key portions of input data, particularly in scenarios with limited training samples. Inspired by human perception, we propose that focusing on essential data segments, rather than the entire dataset, can improve the accuracy and reliability of the learning models. However, identifying these critical data segments, or "hard attention finding," is challenging, especially in few-shot learning, due to the scarcity of training data and the complexity of model parameters. To address this, we introduce LaHA, a novel framework that leverages language-guided deep reinforcement learning to identify and utilize informative data regions, thereby improving both interpretability and performance. Extensive experiments on benchmark datasets validate the effectiveness of LaHA.}
\end{abstract}    
\section{Introduction}
\label{sec:intro}
Few-shot Learning (FSL) has significant applications across various fields \b{such as }image classification \cite{dhillon2019baseline}, activity recognition \cite{kumar2019protogan}, image retrieval \cite{wang2017few}, object tracking \cite{majee2021few}, as well as in medical fields such as drug discovery \cite{vella2022few}. The objective is to enable a model \b{to learn} with only a few training examples per class.
\b{Early FSL methods rely on metric-based and meta-learning approaches. Metric-based methods, such as Prototypical Networks (ProtoNet) \cite{snell2017prototypical}, use class prototypes, which are representative feature vectors for each class, to cluster similar examples in the feature space, enabling classification by finding the nearest prototype for each query. In contrast, meta-learning approaches like Model-Agnostic Meta-Learning (MAML) \cite{finn2017model} focus on optimizing model initialization to facilitate fast adaptation to new tasks.}
Recent advancements in FSL incorporate transductive learning, which leverages both labeled support sets and unlabeled query sets to refine classification. Some popular methods include LaplacianShot \cite{ziko2020laplacian}, which refines class prototypes using a graph Laplacian, and ProtoLP \cite{zhu2023transductive}, which adjusts a bipartite graph to address class imbalance. HyperShot \cite{sendera2023hypershot} introduces kernel-based hypernetworks to dynamically generate task-specific classifiers, enabling robust few-shot learning without gradient-based adaptation.
\s{Additional related methods are detailed in the Appendix.}
\s{\begin{figure}[t]
\centerline{\includegraphics[width=0.45\linewidth]{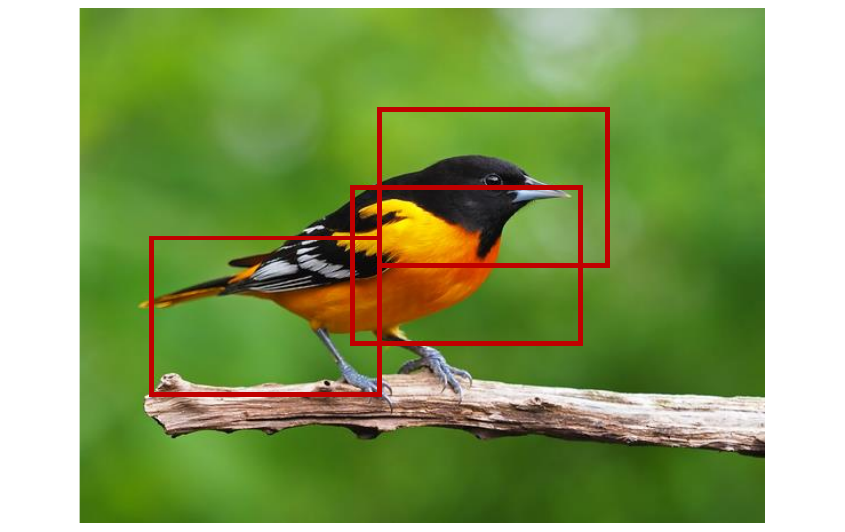}}
\caption{The three attentive regions to define a bird.}
\label{fig:rab}
\end{figure}}


Large Language Models (LLMs) have shown significant potential in FSL by leveraging prior knowledge to address data scarcity. VLM-guided Explicit-Implicit Complementary (VEIC) \cite{zhao2024vlm} enhances few-shot object detection by extracting explicit examples (clearly visible novel class instances) and implicit examples (context-dependent representations), using background augmentation to prevent overfitting and improve robustness. LLaFS \cite{zhu2024llafs} advances few-shot segmentation by integrating task instructions, in-context visual guidance, and curriculum pretraining, where the model is first trained on simpler synthetic data before gradually adapting to real-world data. These strategies enable LLaFS to achieve state-of-the-art performance. Additional related methods are detailed in the Appendix.

\brown{Attention mechanisms, inspired by the human visual system’s ability to focus on key regions, are widely used in computer vision to improve accuracy and interpretability \cite{liu2019end}. Broadly, attention is categorized into soft and hard attention. Soft attention assigns continuous weights (ranging between 0 and 1) to different parts of the input, making it fully differentiable and easy to train with gradient-based optimization \cite{lv2023ssagcn, kocabas2021pare}. Its efficiency, stability, and compatibility with standard deep learning architectures have contributed to its widespread use.}
\brown{Hard attention, in contrast, selects discrete regions of the input, assigning binary weights (0 or 1) \cite{mnih2014recurrent, nikpour2023spatio, nikpour2023spatial}, which provides unique benefits in specific applications. By focusing exclusively on the most informative regions, hard attention reduces redundancy, mitigates noise, and improves interpretability, which makes it ideal for edge devices and real-time applications, reducing computation and memory use by processing only relevant input. Also, by filtering out irrelevant data, hard attention improves the robustness of the model against background variations, lighting changes, and occlusions \cite{ranzato2014learning,alexe2012searching,ba2015learning,seifi2021glimpse,rangrej2021probabilistic}. This is particularly beneficial in few-shot learning, where limited data increases the risk of overfitting to irrelevant details. Unlike models trained on abundant data, few-shot models struggle with high intra-class variation and complex backgrounds. Hard attention addresses this by focusing on task-relevant features, improving generalization. However, the non-differentiable nature of hard attention complicates standard optimization, necessitating alternative techniques such as reinforcement learning (RL). RL treats attention selection as a decision-making process, optimizing it by rewarding the selection of the most informative data regions\cite{mnih2013playing,yun2017action,nikpour2022deep}.}

\brown{While VLMs enhance interpretability by providing semantic insights, their role in task-specific hard attention localization remains underexplored. For example, PaLIGemma detects entities using bounding boxes, making it effective for static object detection \cite{beyer2024paligemma}, but it lacks adaptability in identifying fine-grained regions relevant to downstream tasks like FSL. Though VLMs can be trained for specific objectives, they do not inherently determine the most relevant sub-object regions for FSL or other specialized tasks. This limitation can be addressed by RL, which allows the model to identify and select task-relevant regions, ensuring attention is optimized for the task instead of relying only on predefined object detection methods like bounding boxes. Previous RL-based methods for attention finding in FSL, such as Reinforced Attention Policy (RAP) \cite{hong2021reinforced}, focus on soft attention over feature map of predefined layers rather than the original image. These gaps highlight the need for a method that effectively identifies hard attention and enhances verifiable decision-making directly on the original images in few-shot learning scenarios.}

    \b{Motivated by the underexplored potential of hard attention in few-shot learning, we introduce LaHA (Language-guided Reinforcement Learning for Hard Attention), which uses reinforcement learning to identify critical regions in images for FSL. LaHA frames this non-differentiable task as a Markov Decision Process (MDP), enabling systematic localization of informative image patches using a Vision Transformer (ViT) as the agent. The ViT agent is guided by feedback from a few-shot baseline classifier and a VLM, improving classification accuracy and enhancing interpretability. Also, LaHA constructs a graph of informative patches and uses it as input to the baseline classifier, capturing spatial and contextual relationships. To further improve training, we integrate a contrastive learning task, aiding the model in learning more effective data representations. Despite the typical need for large datasets with ViTs, LaHA’s RL framework and contrastive learning make it robust and adaptable for data-limited regimes. Also, unlike predefined methods using bounding boxes or masks, LaHA captures subtle, task-relevant features. The RL framework optimizes selection to balance accuracy, interpretability, and efficiency, offering a flexible and adaptive alternative to static methods.  To our knowledge, it is the first approach to explore hard attention on input RGB images in few-shot learning, unlike methods that rely on feature maps from transformations. Moreover, existing hard attention methods for classification are not tailored for few-shot settings, often lacking adaptability to limited data. The contributions of the proposed LaHA are summarized below:}
\b{\begin{itemize} \item We introduce the novel problem of hard attention finding for RGB images in few-shot learning. \item We propose a reinforcement learning-based framework using a ViT agent and a graph of attentive regions to preserve spatial information. \item We integrate a vision-language model and contrastive learning to improve human interpretability and training, achieving competitive results on four datasets with limited data. \item Our method reduces data size and storage needs, making it suitable for edge devices and data-sensitive environments. \item We demonstrate the method’s effectiveness in both few-shot learning and general classification, highlighting its versatility. \end{itemize}}

\s{\brown{\b{Motivated by the need for efficient attention mechanisms}\s{Motivated by that, in this paper}, we present LaHA (Language-guided reinforcement learning for Hard Attention), which leverages RL to identify critical regions in images for few-shot learning. 
By framing this non-differentiable task as a Markov Decision Process (MDP), LaHA systematically locates informative image patches within an input image using a Vision Transformer (ViT) as the agent. The ViT agent is guided by feedback from a few-shot baseline classifier and a VLM, enhancing classification accuracy and leveraging the VLM’s human-language basis for interpretability. Additionally, \r{constructing a graph on the informative patches} enables LaHA to capture the spatial and contextual relationships between patches. To further improve training, we incorporate contrastive learning as an auxiliary task to help the model learn more effective data representations. Although ViTs usually require large datasets, LaHA’s RL framework along with the contrastive learning task enhances its robustness and adaptability, making it effective in data-limited scenarios like medical imaging. Also, LaHA operates directly on the original image, identifying hard attention regions that are verifiable through visual inspection, thus making the model's decisions interpretable. LaHA is the first approach, to the best of our knowledge, that explores hard attention for RGB images in the context of few-shot learning. Additionally, existing hard attention methods used in classification tasks tend to have complex structures, making them unsuitable for few-shot setups. The contributions of our method are \b{summarized below}\s{as follows}:}}
\brown{ \s{\r{TERRIBLE WRITING for this paragraph! This is too details, and more like an overview of the method. } Inspired by these benefits, we propose LaHA (Language-guided reinforcement learning for Hard Attention), a novel method for detecting hard attention regions in few-shot learning. LaHA formulates the \b{non-differentiable} task of finding attentive regions as a Markov Decision Process (MDP). This approach allows LaHA to systematically explore \b{the input image} and select the most informative \b{image patches}\s{ within an image}. We employ a ViT as the RL agent, taking advantage of its ability to capture contextual information in images more effectively than traditional convolutional architectures. The agent identifies the optimal locations of multiple attentive areas (patches) within an image. The reward for the agent comes from two sources: (1) feedback from the few-shot baseline classifier, which shows how well the selected attentive regions contribute to the overall classification task, and (2) a reward from a VLM that enhances interpretability by ensuring that the patches align with the image's semantic content. After identifying these attentive patches, we model the selected regions as a graph, where each patch is considered a node, and the relationships between these patches are represented by an adjacency matrix. This graph-based modeling approach enables us to capture the spatial and contextual relationships between different regions within the image, further enhancing the model’s understanding of the underlying structure. To further improve training, we incorporate contrastive learning as an auxiliary task to help the model learn more effective data representations. Although ViTs generally benefit from large datasets, LaHA overcomes this limitation by using hard attention to focus on key informative regions within an image, enabling it to capture essential contextual information, reducing its dependency on large datasets. Additionally, LaHA’s auxiliary contrastive learning task helps the model learn robust representations from minimal data, enhancing its adaptability and generalization in few-shot learning scenarios.  }}
\s{Deep RL has been successfully applied to various computer vision tasks \cite{mnih2013playing,yun2017action,nikpour2023spatial,nikpour2021joint,nikpour2022deep}, including methods aimed at finding hard attention in RGB images for classification tasks \cite{ba2014multiple,elsayed2019saccader}. Previous methods that used reinforcement learning to find soft attention in few-shot learning, such as the Reinforced-Attention Policy (RAP) \cite{hong2021reinforced}, focused on the feature maps of a predefined layer rather than the original image, which limited their ability to reduce image size and undermined interpretability. LaHA, in contrast, operates directly on the original image, identifying hard attention regions that are verifiable through visual inspection, thus making the model's decisions more interpretable. LaHA is the first approach, to the best of our knowledge, that explores hard attention for RGB images in the context of few-shot learning. Additionally, existing hard attention methods used in classification tasks tend to have complex structures, making them unsuitable for few-shot setups. The contributions of our method are \b{summarized below}\s{ as follows}:}
\s{%
\brown{\begin{itemize}
    \item We introduce the novel problem of hard attention finding for RGB images in few-shot learning setups.
    \item We propose a reinforcement learning-based framework that uses a ViT agent and constructs a graph of attentive regions to maintain spatial information.
    \item To help human interpretable selection and improve training, we integrate a vision-language model and a contrastive learning module, achieving competitive results on four datasets while using only a small portion of the data.
    \item By focusing on informative regions, our method reduces data size, and storage requirements, making it suitable for edge devices and data-sensitive environments.
    \item We demonstrate the effectiveness of our method not only in few-shot classification tasks but also in the general classification, proving its versatility.
\end{itemize}}}

\section{Methodology}
In few-shot learning, an \textit{episode}\footnote{Two types of episodes are used in this paper: the episode in the reinforcement learning framework and the \textit{episode} in the few-shot setup, which is defined by italics to avoid confusion.} refers to a single learning task where a model \b{must} learn from a limited amount of data. \b{The few-shot model is trained across many \textit{episodes}. Each \textit{episode} typically consists of two main components: a Support set and a Query set. To train the model, a support set and a query set are randomly selected from the data in each \textit{episode}. The classes in both sets are the same, but the individual samples within the sets do not overlap. This learning approach is commonly referred to as ``$N$-way, $G$-shot'' classification, where $N$ denotes the number of classes and $G$ represents the number of samples per class in both the support and query sets. We define the support set as $SU = {(x_j, y_j)}_{j = 1}^{N \times G}$, where $x_j$ is the $j^{th}$ sample in the support set and $y_j$ is its corresponding label.}
\s{In few-shot learning, an \textit{episode}\footnote{Two types of episodes are used in this paper: the episode in the reinforcement learning framework, and the \textit{episode} in the few-shot setup which is defined by italic word to avoid confusion.} refers to a single learning task where a model has to learn from a limited amount of data. The few-shot model learns during many \textit{episodes}. Typically, an \textit{episode} consists of two main components: a Support set and a Query set. To train a model, a support set and a query set are randomly selected from the data in each \textit{episode}. This approach of learning is commonly referred to as ``$N$-way, $G$-shot'' classification, where there are $N$ classes and $G$ samples for each class in the support and query sets. We denote the support set as $SU = \{(x_j, y_j)\}_{j = 1}^{N\times G}$, where $x_j$ is the $j^{th}$ sample in the support set and $y_j$ is its corresponding label.} 

\s{\subsection{LaHA Framework}\r{or overview? this section and the one before, do not have to be under a separate section. You may merge them and have them more like your opening paragraphs. Then have three subsections: RL module, Baseline Classifier module, Contrastive learning module. and then a closing paragraph talking a bout the total loss without having a separate section for it.}}
\b{In our proposed method, we assume that each object can be recognized by a set of $N_a$ key regions (also called attentive regions or patches). LaHA aims to detect $N_a$ attentive image patches, which may overlap. By allowing overlap, LaHA can handle objects at different scales within the image. Each patch has a fixed size, set to $\lfloor \frac{d_1}{N_a} \rfloor \times \lfloor \frac{d_2}{N_a} \rfloor$, where $\lfloor . \rfloor$ denotes the floor function, and $d_1$ and $d_2$ refer to the image's width and height, respectively. This ensures that LaHA covers a maximum of $1/N_a$ and a minimum of $1/N_a^2$ of the input image. 
We set $N_a = 3$ in this sections as it yielded the best performance in most experiments. Additional evaluations on different patch sizes and numbers are provided in the Appendix.}
 \begin{figure}[t] 
    \centering    
  \includegraphics[width=0.5\textwidth]{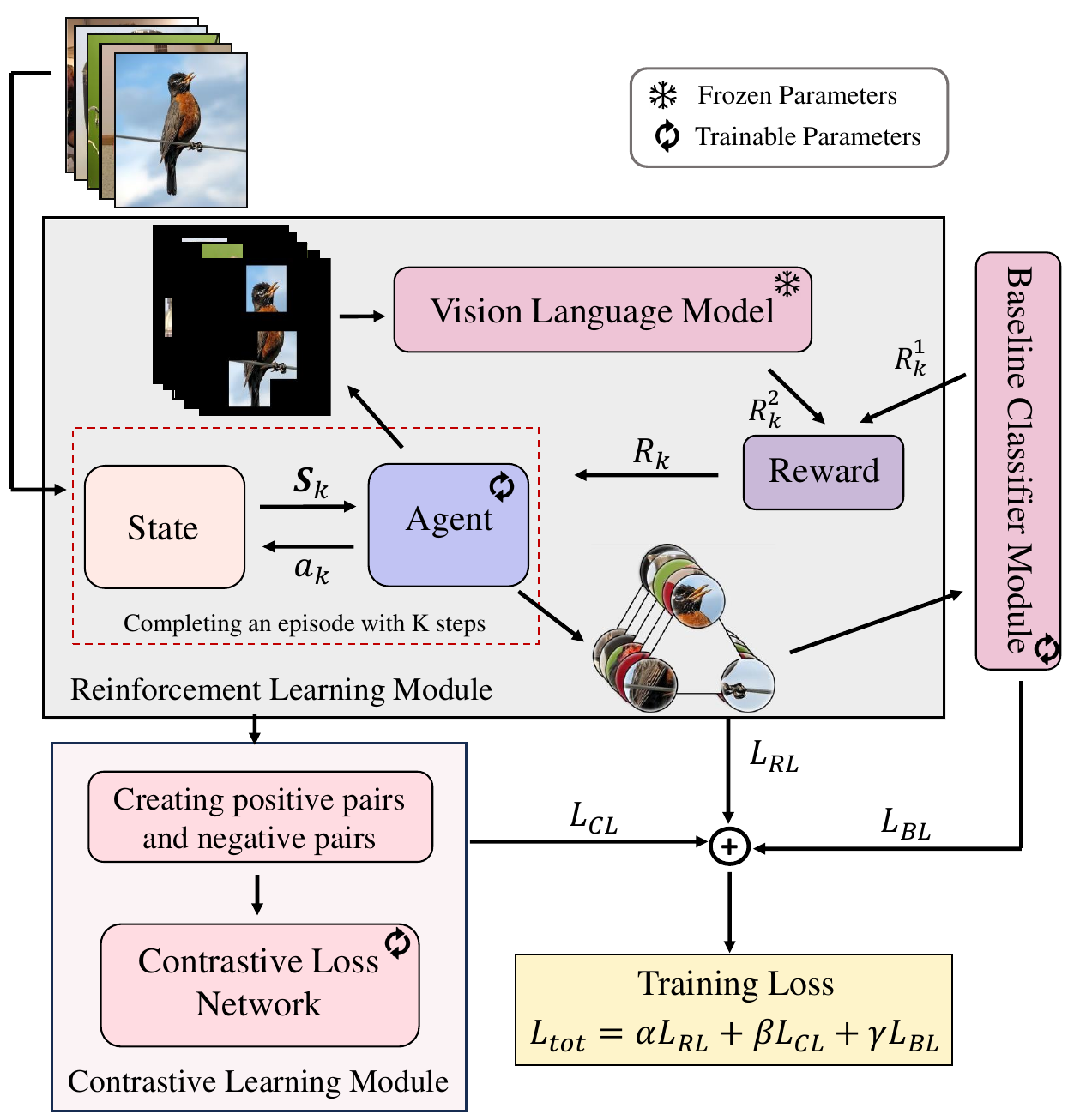}
    \caption{Block diagram of the proposed LaHA \b{framework}.}
    \label{fig:bd}
\end{figure}

\b{The block diagram of the proposed LaHA framework is shown in Figure \ref{fig:bd}. LaHA consists of three main modules: (1) an RL module responsible for identifying the attentive regions, (2) a baseline classifier module for performing the final classification task, and (3) a contrastive learning module, which is added as an auxiliary task to enhance model's performance and generalization.
In the RL module, the state is defined using the input data and is processed by an agent. The agent’s output consists of the three selected regions (attentive patches) per image. These patches are used to construct a graph representing the spatial and contextual relationships between the selected regions, which is then provided as input to the FSL baseline classifier to compute the classification loss, denoted as $L_{BL}$. Additionally, the agent's output is used to generate positive and negative pairs, which are employed to calculate the contrastive loss, denoted as $L_{CL}$.
The agent executes an episode consisting of $K$ steps, receives a reward, and calculates the RL loss, denoted as $L_{RL}$. The reward is computed based on feedback from two sources: the baseline classifier and a VLM. 
Finally, the three losses, $L_{RL}$, $L_{BL}$, and $L_{CL}$, are combined with different weights to form the total loss, $L_{tot}$, which is used to update the networks of all three modules. The details of each module are discussed below.}

\s{The block diagram of the proposed LaHA framework is shown in Figure \ref{fig:bd}. LaHA consists of three main modules: an RL module, which is responsible for finding the attentive regions, the baseline classifier module, for learning the final classification task, and a contrastive learning module which we add as an auxiliary task to improve the performance and generalization of the model. 
In short, in the RL module, the state is defined using the data and is processed by an agent. \b{The agent's output is the three selected regions per image. The patches of every image are used to form a graph and the graph of each image is given to the FSL baseline classifier to compute the classifier loss,  i.e., $L_{BL}$.} 
The agent completes an episode of $K$ steps, gets a reward, and calculates the loss of the RL module, i.e., $L_{RL}$.
The reward is generated from two sources: the baseline classifier and a VLM. Also, the agent output's on the training data is used to create positive and negative pairs to be used to calculate the contrastive loss, i.e. $L_{CL}$. In the end, these three losses are added with different weights to generate the total loss, $L_{tot}$, used to update the networks of all three modules. The details regarding each module are discussed below.}

\subsection{Reinforcement Learning Module}
\begin{figure}[t]
\centerline{\includegraphics[width=0.35\linewidth]{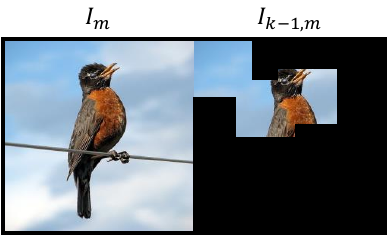}}
\caption{The agent's state $S_{k,m}$, which is concatenation of the original image $I_m$, and the image output of the agent in the $k-1^{th}$ step of the episode $I_{k-1,m}$, i.e. $S_{k,m} = [I_m, I_{k-1,m}]$. }
\label{fig:state}
\end{figure}

\begin{figure}[ht]
\centerline{\includegraphics[width=0.72\linewidth]{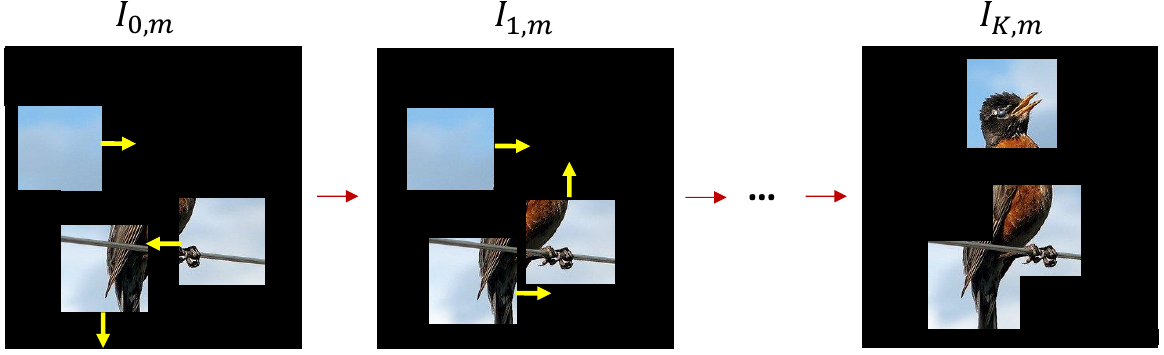}}
\caption{An example of $I_{k,m}$, $\{k=0,..,K\}$ during one episode in our proposed method. The actions, shown in yellow arrows, for step 1 are go right ($\rightarrow$), go down ($\downarrow$), and go left ($\leftarrow$) for the first, second, and third regions respectively, which results in $I_{1,m}$. After completing an episode, the agent outputs the regions found in $I_{K,m}$.}
\label{fig:eps}
\end{figure}

In this paper, the RL agent is run for \b{every image in the batch, with batch size denoted by $ M$. The $k^{th}$ step} in the RL episode for the $m^{th}$ sample in the batch is represented as $\mathcal{T}_{k,m}=(S_{k,m},A_{k,m}, R_{k})$, where $S_{k,m}$, $A_{k,m}$, and $R_{k}$ denote the state, action, and reward at the $k^{th}$ step for the $m^{th}$ image in the batch. \b{A complete episode for the $m^{th}$ sample is denoted by} $\mathcal{T}_{m}=(S_{1,m},A_{1,m}, R_{1}, ..., S_{K,m},A_{K,m}, R_{K})$. \b{State, action, reward and agent are defined as follows}:
\\
\textbf{State:} We define the state \( S_{k,m} \), for \( m = 1, \dots, M \), as the concatenation of the \( m^{\text{th}} \) original image in the batch, \( I_m \), and the agent's output from the previous step, \( I_{k-1,m} \), i.e., \( S_{k,m} = [I_m, I_{k-1,m}] \). Here, \( I_{k-1,m} \) has the same dimensions as \( I_m \) but only retains the selected patches, with all non-selected regions set to zero. An example of \( I_{k-1,m} \) is shown on the right side of Figure \ref{fig:state}. For the initial state \( S_{0,m} \), \( I_{0,m} \) is initialized with patches randomly located across the image, as illustrated in Figure \ref{fig:eps}.
\\
\textbf{Action:} \b{At each step of the episode, specifically the \( k^{\text{th}} \) step for the \( m^{\text{th}} \) image, the agent outputs a probability matrix \( P_{k,m} \) through its policy head. This probability matrix is used to determine the agent's actions, which define the movement direction for each patch. Five possible actions are designed for each patch: move up, down, left, right, or remain in place, with all movements occurring at a step size \( b \). For the \( m^{\text{th}} \) image, each action is denoted by \( a_{k,m}^l \), where \( k \) indicates the current step of the RL episode and \( l \in \{1, 2, \ldots, N_a\} \) specifies the \( l^{\text{th}} \) attentive region. The actions are sampled from a categorical distribution formed based on the agent's probability output \( P_{k,m} \). An example of one episode is illustrated in Figure \ref{fig:eps}.}\s{At each step of the episode, i.e. the $k^{th}$ step, and for the $m^{th}$ image, the agent outputs the probability matrix $P_{k,m}$, through its policy head, which will be used to define the actions it should take. The action defines the direction of movements of the patches. Therefore, for each patch, five actions have been designed: go up, down, left, right, and apply no change. All the movements are done with step size $b$. For the $m^{th}$ image, each action is denoted by $a_{k,m}^l$ where $k$ refers to the $k^{th}$ step of the RL episode and $l \in \{1, 2, 3\}$ shows the $l^{th}$ attentive region. The actions are sampled from a categorical distribution formed based on the probability output of the agent, $P_{k,m}$. An example of one episode is depicted in Figure \ref{fig:eps}.}
\\
\textbf{Reward}:
\b{The reward function should reflect how well the agent’s action aligns with its goal. Our method aims to enhance the baseline classifier's performance and add human-level semantic interpretability, aligning selected regions with human perception. Thus, we define two reward components: (1) a classification-based reward from the FSL baseline classifier, denoted by \( R_k^1 \), and (2) a VLM-based reward to enhance interpretability, denoted by \( R_k^2 \). To compute \( R_k^1 \), the baseline classifier evaluates the agent's policy at step \( k \) of the episode by applying it to the batch of input images, which contains \( N \times G \) samples from both support and query sets. For each image, the \( N_a \) selected regions form nodes in a graph, with edges representing the relationships between these regions based on the cosine similarity of their feature vectors (details on graph construction are provided in Section \ref{Training the Baseline model}). This approach ensures that the graph captures semantic relationships between patches. The batch of these constructed graphs is then fed into the few-shot baseline classifier to calculate the classification loss, and the average loss across the batch is taken as the negative of \( R_k^1 \), forming the classification-based reward.
\\
\( R_k^2 \) aims to enhance interpretability. To achieve this, we mask (i.e., zero out) the unselected parts of each image in the batch and feed the modified images into a VLM, using a prompt that directs it to describe the selected regions (an example prompt is shown in Figure \ref{molmo:pd}). We then compare the VLM’s description of the modified image to its description of the original image using BERTScore, which quantifies semantic similarity between the two outputs \cite{zhang2020bertscore}. This reward, computed as the average BERTScore across the batch, ensures that the selected regions are informative for classification and semantically consistent with the full image content. The agent's total reward is calculated as:}
\s{The reward should reflect how good the action of the agent is in reaching its goal. The goal of our method is to improve the baseline \b{classifier} performance and \brown{add human-level semantic info, leading to obtaining selected regions aligned with human interpretation.} Therefore, we define two reward terms: (1) a classification-based reward from \b{a} few-shot baseline classifier \b{denoted by $R_k^1$}, and (2) a \s{Vision-Language Model }VLM-based reward to enhance interpretability, \b{denoted by $R_k^2$}. \b{Towards finding} $R_k^1$, the baseline \b{classifier} evaluates the \brown{performance of the agent's policy} at step $k$ of the episode by applying it to the \b{images of the input batch}\s{samples batch}, containing $N\times G$ samples from both the support and query sets. Then we use the \b{$N_a$} selected regions of each image to construct a graph\b{, where each selected patch of the image} represents a node, and the relationships between the regions are captured by edges based on the cosine similarity of their \brown{feature vectors \b{obtained through a CNN network} (further details on graph construction are provided in Section \ref{Training the Baseline model})}.\s{Using cosine similarity ensures the graph captures semantic relationships between patches, focusing on content similarity.} \brown{The batch of these constructed graphs, derived from the input data batch, is then fed into the baseline classifier to calculate the loss. The average loss across the batch is taken as the negative of $R_k^1$, forming the classification-based reward.}
\s{For the second reward, } $R_k^2$ \b{aims} at improving interpretability. \b{To this end,} we mask \b{(i.e. zero-out)} the unselected parts of each image in the batch and pass the modified image to a VLM, \brown{ providing a prompt that guides it to describe the selected regions. An example of a prompt is shown in \r{Figure \ref{molmo:pd}}}. The VLM's output on the modified image is then compared to its output on the original image using the BERTScore, which quantifies the semantic similarity between the two descriptions \cite{zhang2020bertscore}. This second reward, obtained by averaging the BERTScore over the batch, ensures that the selected regions are not only informative for classification but also semantically aligned with the overall content of the image. The total reward of the agent \b{is shown below}:}
\begin{equation}
\label{eq_rew}
R_k = \xi_1 R^1_k + \xi_2 R^2_k,  
\end{equation} 
where $\xi_1$ and $\xi_2$ are coefficients that control the balance between the contributions of their corresponding terms. \s{\brown{\r{IS THIS Needed here? You may move it to appendix. It just is off the current trend of the paper. It is just a tiny implementation detail??} \r{It should be noted that we used a random batch of the validation set to calculate both of the rewards to increase the method's generalization ability.}}}
\\
\brown{\b{\textbf{Agent:}} We integrate a ViT as the agent's backbone to extract hierarchical spatial features from its input state, enabling it to capture both local and global dependencies in visual data. These features are then processed by a policy head, which outputs action probabilities \( P_{k,m} \), and a value head, which estimates the value of the current state. This design leverages ViT’s feature extraction capabilities to enhance decision-making within the RL framework.}
\\\\
\textbf{\brown{Reinforcement Learning Loss:}} We use Proximal Policy Optimization (PPO) in our method for its efficient and stable policy updates in reinforcement learning \cite{schulman2017proximal}. PPO applies a clipping function to control the policy update magnitude, limiting the change in probability ratio between the new and old policies. This method prevents large shifts in policy, reducing variance and improving convergence\, making it well-suited to our task\s{ by enhancing stability and performance}.
\\
At each step $k$, the total reward $R_k$ is calculated over the \b{input images} batch. The advantage estimate $\hat{A}_k$ helps quantify how much better (or worse) an action was compared to the expected value of being in a particular state. The advantage is computed as:
\begin{equation}
\begin{split}
\label{ad}
\hat{A}_{k,m} = R_k - \frac{1}{M} \sum_{m=1}^M V(S_{k,m})
\end{split}
\end{equation}
where $R_k$ is the total reward at step $k$ and $V(S_{k,m})$ is the value function, which estimates the expected return for the state $S_{k,m}$. 
In PPO, the policy\s{, parameterized by $\theta^p$,} is updated by maximizing a clipped objective function. To ensure stable updates, the probability ratio $r_k(\theta)$ between the new and old policies is constrained within a certain range. The policy's loss function, $L_{P}$, is:
\begin{multline}
L_{P} = \frac{1}{KM} \sum_{k=1}^K \sum_{m=1}^M \min \Big( r_{k,m}(\theta) \hat{A}_{k,m}, \\
\text{clip}\left( r_{k,m}(\theta), 1 - \epsilon, 1 + \epsilon \right) \hat{A}_{k,m} \Big) + \iota H(\pi_{\theta}(\cdot | S_{k,m})),
\end{multline}
\begin{equation}
r_{k,m}(\theta) = \frac{\pi_{\theta}(a_{k,m} | S_{k,m})}{\pi_{\theta_{\text{old}}}(a_{k,m} | S_{k,m})}
\end{equation}
where $\pi_{\theta(.)}$ and $\pi_{\theta_{\text{old}}}(.)$ represent the current and the previous policies, respectively, and $H(.)$, controlled by coefficient $\iota$, is the entropy term to add exploration, defined as:  
\begin{multline}
H(\pi_{\theta}(\cdot \mid S_{k,m})) = - \sum_{a_{k,m}} (\pi_{\theta}(a_{k,m} \mid S_{k,m})) \\
(\log \pi_{\theta}(a_{k,m} \mid S_{k,m}))
\end{multline}
The value network is updated by minimizing the mean squared error between the predicted value \( V(S_{k,m};\theta) \) and the observed return \( R_k \). The value loss function, $L_V$, is given by:
\begin{equation}
L_V = \frac{1}{KM} \sum_{k=1}^K \sum_{m=1}^M \left( V(S_{k,m};\theta) - R_k \right)^2
\end{equation}
where \( K \) is the number of steps in the episode, and \( M \) is the batch size. For simplicity, $\theta$ represents the parameters of both the policy and value networks. \s{This loss function encourages the value network to improve its predictions of expected returns for each state \( S_{k,m} \), aiding in more accurate advantage calculations for policy updates.}
The total RL loss, $L_{RL}$, combines the policy and value losses, weighted by coefficients $\eta$ and $\mu$ as:
\begin{equation} L_{RL} = \eta L_{P} + \mu L_V \end{equation}
This \brown{total RL loss, $L_{RL}$,} is then used to update ViT (the shared network), with gradients from both the PPO policy and value losses contributing to the parameter updates.

\subsection{Baseline Module}\label{Training the Baseline model} 
\b{The baseline module is responsible for classifying few-shot samples using only the selected attentive regions identified by the agent. After the agent finds these regions, we construct an undirected graph for each image, where each selected region is first passed through a lightweight shared convolutional layer to extract feature embedding serving as node representation. The nodes in each image are all linked, forming a complete graph. Edge weights are computed using cosine similarity between node embeddings, capturing semantic relationships, and are normalized for stable training. The graph remains static to maintain consistency in representation learning. Finally, a Graph Neural Network (GNN) processes the graph, refining both node-level information (i.e., the characteristics of each individual patch) and graph-level information (i.e., the overall structure and relationships between patches through the connections between nodes). The output from the GNN is then passed into a FSL classification baseline model, \grn{which classifies the samples based on the graph-structured information and calculates its corresponding loss, considered as baseline module's loss, $L_{BL}$}. This loss function helps differentiate the classes effectively in a few-shot setting by leveraging the graph-based features extracted from the attentive regions.}
\s{The baseline module is responsible for the primary task of classifying few-shot samples using only the selected attentive regions identified by the agent. First, after obtaining these attentive regions as the output of the agent, we construct an undirected graph for each sample where the selected regions serve as nodes. In this graph, all nodes are connected to each other, forming a fully connected (complete) graph. \brown{To capture the semantic relationship between the patches, } the edges between nodes are defined using cosine similarity of their feature representations. \brown{These representations are obtained by passing the selected regions to a convolutional layer}.}
\s{Next, to extract meaningful representations from this graph, we apply a Graph Neural Network (GNN). \brown{The GNN processes the fully connected undirected graph to learn both node-level information, i.e. the characteristics of each individual patch and graph-level information, i.e. overall structure and relationships between patches through the connections between nodes.} Then, the output from the GNN is fed into a FSL classification baseline model, \grn{which classifies the samples based on the graph-structured information and calculates its corresponding loss, considered as baseline module's loss, $L_{BL}$.}}
\begin{figure}[b]
\centerline{\includegraphics[width=0.77\linewidth]{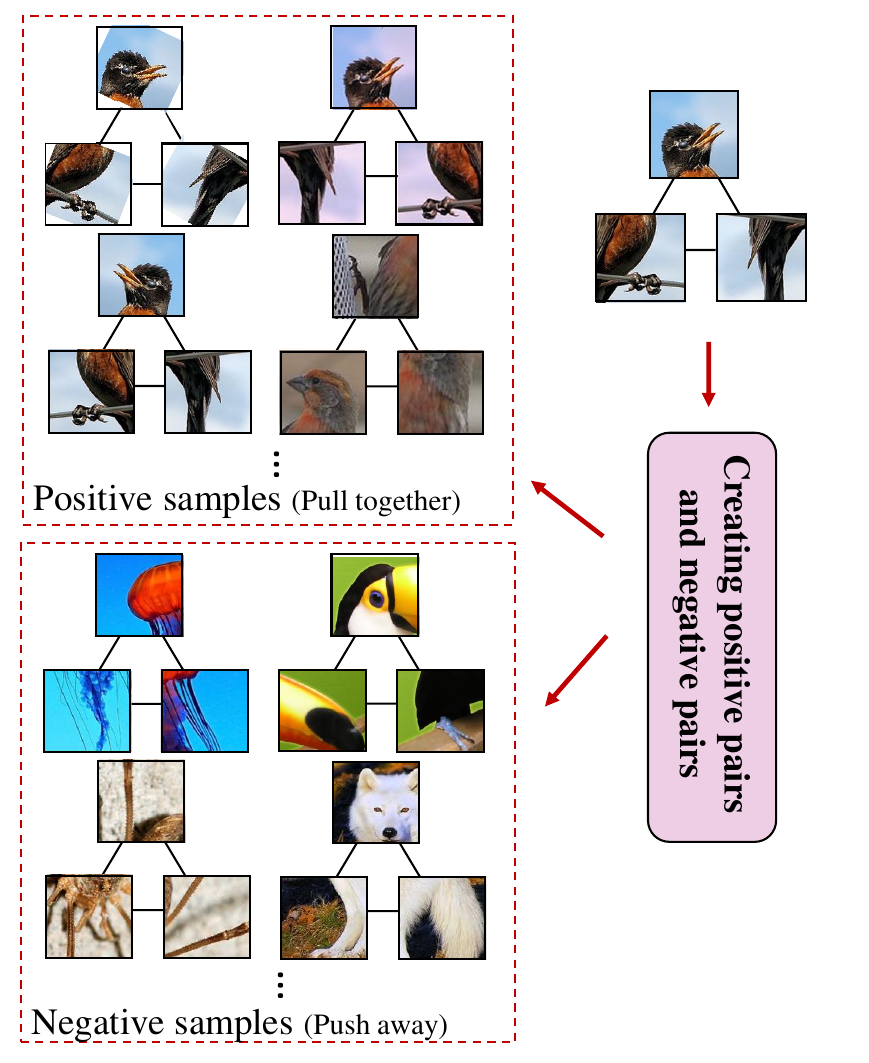}}
\caption{An example of creating positive and negative pairs in 5-way scenario.}
\label{fig:aug}
\end{figure}
\begin{table*}[ht!]
    \centering
        \caption{\b{Few-shot classification} performance comparison across different methods and datasets for few-shot learning. Results are reported as mean $\pm$ standard deviation for each setting. \b{MOLMO is used in LaHA. }}
    \resizebox{\textwidth}{!}{%
    \begin{tabular}
    {c|l|c|c|c|c|c|c|c}
        \hline
        \multirow{2}{*}{\textbf{Methods}} & \multicolumn{2}{c|}{\textbf{MiniImageNet}} & \multicolumn{2}{c|}{\textbf{CIFAR-FS}} & \multicolumn{2}{c|}{\textbf{FC-100}} & \multicolumn{2}{c}{\textbf{CUB}} \\ \cline{2-9}
         & \textbf{5-s 5-w} & \textbf{1-s 5-w} & \textbf{5-s 5-w} & \textbf{1-s 5-w} & \textbf{5-s 5-w} & \textbf{1-s 5-w} & \textbf{5-s 5-w} & \textbf{1-s 5-w} \\ \hline
        ProtoNet (ResNet-18) \cite{snell2017prototypical} & 73.68 $\pm$ 0.65 & 54.19 $\pm$ 0.82 & 80.12 $\pm$ 0.65 & 68.15 $\pm$ 0.60 & 51.12 $\pm$ 0.85 & 36.54 $\pm$ 0.80 & 82.54 $\pm$ 0.90 & 68.15 $\pm$ 0.85 \\ 
        LaplacianShot (WRN) \cite{ziko2020laplacian} & 85.12  $\pm$ 0.14 & 74.86 $\pm$ 0.19 & 85.34 $\pm$ 0.43 & 75.23 $\pm$ 0.39 & 55.87 $\pm$ 0.65 & 39.32 $\pm$ 0.69 & 88.68 $\pm$ 0.69 & 80.96 $\pm$ 0.69 \\ 
        HyperShot (ResNet-10) \cite{sendera2023hypershot} & 69.62 $\pm$ 0.20 & 53.18 $\pm$ 0.50 & 78.20 $\pm$ 0.45 & 60.15 $\pm$ 0.55 & 50.30 $\pm$ 0.70 & 34.20 $\pm$ 0.60 & 80.07 $\pm$ 0.22 & 66.13 $\pm$ 0.26 \\ 
        protoLP (ResNet-12)\cite{zhu2023transductive} & 90.22 $\pm$ 0.11 & 84.35 $\pm$ 0.24 &  91.52 $\pm$ 0.15 &  88.22 $\pm$ 0.21 & 57.50 $\pm$ 0.65 & 41.00 $\pm$ 0.70 & 94.65 $\pm$ 0.10 &  91.82$\pm$ 0.18  \\ \hline
        LaHA\_ProtoNet (ResNet-18) & \textbf{79.54 $\pm$ 0.13} & \textbf{61.29 $\pm$ 0.60} & \textbf{82.47 $\pm$ 0.65} & \textbf{71.02 $\pm$ 0.60} & \textbf{53.32 $\pm$ 0.80} & \textbf{39.00 $\pm$ 0.75} & \textbf{90.77 $\pm$ 0.41} & \textbf{75.59 $\pm$ 0.43}  \\ 
        LaHA\_LaplacianShot (WRN) & \textbf{89.12 $\pm$ 0.24} & \textbf{82.02 $\pm$ 0.41} & \textbf{87.00 $\pm$ 0.44} & \textbf{75.80 $\pm$ 0.45} & \textbf{58.00 $\pm$ 0.70} & \textbf{43.00 $\pm$ 0.68} & \textbf{93.12 $\pm$ 0.25} & \textbf{86.02 $\pm$ 0.41} \\
        LaHA\_HyperShot (ResNet-10) & \textbf{72.37 $\pm$ 0.26} & \textbf{55.98 $\pm$ 0.63} & \textbf{80.65 $\pm$ 0.60} & \textbf{63.87 $\pm$ 0.55} & \textbf{51.93 $\pm$ 0.65} & \textbf{37.98 $\pm$ 0.70} & \textbf{82.93 $\pm$ 0.31} & \textbf{68.14 $\pm$ 0.35} \\ 
        LaHA\_protoLP (ResNet-12) & \textbf{94.92 $\pm$ 0.22} & \textbf{88.97 $\pm$ 0.31} &
        \textbf{94.98 $\pm$ 0.41} & \textbf{90.12 $\pm$ 0.27} & \textbf{64.82 $\pm$ 0.55} & \textbf{49.67 $\pm$ 0.49} & \textbf{96.21 $\pm$ 0.25} & \textbf{94.22 $\pm$ 0.24} \\ \hline
    \end{tabular}%
    }
\label{tab:results}
\end{table*} 

\subsection{Contrastive Learning Module} 
\brown{Incorporating auxiliary tasks can enhance the main task's performance, improve generalization, and accelerate convergence by providing additional gradients that guide learning. These tasks also encourage the model to learn meaningful, disentangled representations, which aid in feature extraction. To leverage these benefits, we incorporate a contrastive learning module as an auxiliary task to strengthen the training of our LaHA framework.} Contrastive learning is a learning technique that aims to learn useful representations by bringing similar data samples, i.e. positives, closer to each other in the learned representation space while pushing dissimilar data points, i.e. negatives, farther apart. 

In our method, we use the support set and query set in the batch to create positive and negative pairs. Pairs from the same class and their augmentation are considered as positive pairs, while negative pairs are pairs from different classes. For augmentation, (1) We apply a random rotation to the selected patches within the range of -45 degrees to +45 degrees to make our method robust to orientation variations, (2) We horizontally flip the selected patches to encourage the model to become invariant to left-right orientation changes, and (3) We apply color jittering to the selected patches by randomly adjusting brightness, contrast, and saturation within a factor of 0.8 to 1.2, and modifying the hue by ±0.1. This simulates different lighting conditions and enhances the model's robustness to color variations. An example is shown in Figure \ref{fig:aug}.

After augmentation and creating positive and negative graph pairs, the contrastive loss, $L_{CL}$, is calculated. As the labels of data samples are available, we use the supervised contrastive loss (SupCon loss)
\cite{khosla2020supervised} as follows:
\begin{equation}
\begin{split}
\label{eqsup}
L_{CL} = \sum_{i\in W }  \frac{-1}{|O(i)|}\sum_{o\in O(i)}log \frac{\exp (f_i\cdot f_o/\tau)}{\sum_{u\in U(i)}\exp(f_i \cdot f_u/\tau)}
\end{split}
\end{equation}
where $i \in W \equiv \{1, \dots, 2M\}$, \brown{is the index of a single sample in the augmented batch ($M$ is the batch size).} $O(i)$ shows the indices of the positive pairs distinct from $i$, and $|O(i)|$ denotes its cardinality, $U(i) \equiv W \backslash {i}$, $f_\zeta$ shows the output representation of the network for the $\zeta$th index, and $\tau$ is the temperature parameter to control the similarity scale. For more details, see \cite{khosla2020supervised}.

\subsection{\b{Loss Optimization and Inference}}
After calculating the three losses—RL loss, baseline loss, and contrastive learning loss—the total loss to be minimized is computed as:
\begin{equation}
\begin{split}
\label{l_tot}
L_{tot} = \alpha L_{RL} + \beta L_{CL} + \gamma L_{BL}
\end{split}
\end{equation}
where $\alpha$, $\beta$, and $\gamma$ are hyperparameters to control the contribution of their corresponding loss term. \s{\brown{ To enhance training stability and performance, we first partially train the CL module and baseline classifier together before introducing the RL module. This joint training phase provides a robust initial structure for both modules, reducing overfitting and enabling effective adaptation during later stages with the RL module. Specifically, the CL module is trained on random patches from the dataset to develop foundational representations, while the baseline classifier is trained concurrently on these embeddings to minimize classification loss. Once integrated with the RL module, the CL and baseline classifier are updated with lower learning rates for smoother convergence.}\brown{Additionally, a fixed pre-trained VLM is employed to get \b{$R^2_k$ in equation \eqref{eq_rew}. }}} 

\b{The pseudo-code for the proposed LaHA framework is presented in Algorithm 1 of the Appendix. In brief, a batch of few-shot tasks is sampled, and the training data, including both the support and query sets, is provided to the agent. The agent completes $K$ steps, with the reward $R_k$ calculated at each step. The agent outputs the coordinates of the $N_a$ selected patches. Using these coordinates, a new training set is constructed, incorporating graphs of the attentive regions. The baseline loss is then computed based on this new set. The contrastive learning loss is calculated using the output embeddings from the contrastive learning network, applied to both the original data and its augmentation. After calculating the three losses, the total loss is derived, and all the trainable networks are updated. This process is repeated for a predefined number of \textit{episodes} until the three networks converge. \brown{ During training, the RL module, CL module, and FSL baseline with the lightweight GNN are trainable, while the VLM operates in inference mode without being trained. At inference, only the trained policy network, GNN, and FSL baseline remain, meaning the VLM is not required during the test phase. This results in significantly fewer parameters compared to training. Moreover, in some scenarios, the FSL baseline may require retraining without modifying the RL module. For example, in continual learning, where new classes are introduced over time, the RL module’s learned strategy for selecting informative regions remains valid, while the FSL baseline needs to be updated to incorporate new categories. Similarly, if the FSL baseline is deployed in a new dataset with the same type of objects but different categories, retraining it without modifying the RL module can improve generalization. Since LaHA processes only the most informative regions rather than full images, it significantly reduces computational demands in these scenarios while maintaining interpretability and improving performance.} \s{This design balances performance, efficiency, and interpretability.}}
\begin{table*}[ht!]
    \centering
        \caption{Few-shot classification performance comparison on mini-ImageNet, CIFAR-FS, FC-100, and CUB. Results are reported as mean $\pm$ standard deviation.}
    \resizebox{\textwidth}{!}{%
    \begin{tabular}
    {c|l|c|c|c|c|c|c|c}
        \hline
        \multirow{2}{*}{\textbf{Methods}} & \multicolumn{2}{c|}{\textbf{MiniImageNet}} & \multicolumn{2}{c|}{\textbf{CIFAR-FS}} & \multicolumn{2}{c|}{\textbf{FC-100}} & \multicolumn{2}{c}{\textbf{CUB}} \\ \cline{2-9}
         & \textbf{5-s 5-w} & \textbf{1-s 5-w} & \textbf{5-s 5-w} & \textbf{1-s 5-w} & \textbf{5-s 5-w} & \textbf{1-s 5-w} & \textbf{5-s 5-w} & \textbf{1-s 5-w} \\ \hline
         MAML \cite{finn2017model} & 65.72 $\pm$ 0.92 & 49.61 $\pm$ 0.92
         & 80.12 $\pm$ 0.61 & 68.15 $\pm$ 0.59 &51.12 $\pm$ 0.71 & 36.54 $\pm$ 0.52 & 82.59  $\pm$ 0.78 & 70.89 $\pm$ 051 \\
        MatchingNet \cite{vinyals2016matching} & 68.88 $\pm$ 0.69& 52.91 $\pm$ 0.88 & -& -& - & - &-\\
        Cosine classifier \cite{chen2019closer}&-&-&-&-& 57.67  $\pm$ 0.77 & 38.47  $\pm$ 0.70&-&- \\
        TADAM \cite{oreshkin2018tadam} &-&-&-&-&  56.10  $\pm$ 0.40 & 40.10  $\pm$ 0.40 &-&- \\
        SimpleShot \cite{wang2019simpleshot} & 82.09 $\pm$ 0.14 & 65.87 $\pm$ 0.20 & - & - & 53.63 $\pm$ 0.18 & 40.13 $\pm$ 0.18 & 53.63 $\pm$ 0.18 &  40.13 $\pm$ 0.18\\
        SIB\cite{hu2020empirical} & 79.20 $\pm$ 0.40& 70.00 $\pm$ 0.60 & 85.30 $\pm$ 0.40 & 80.00 $\pm$ 0.60 &- &- &- \\
        SSR \cite{shen2021re} &-&-& 83.70 $\pm$ 0.40 & 76.80 $\pm$ 0.60&-&-&-&-\\
        RAP-LaplacianShot \cite{hong2021reinforced} & 75.58 $\pm$ 0.20 & 85.31 $\pm$ 0.08 & - & - & - & - & 90.77  $\pm$ 0.10 & 83.59  $\pm$ 0.18  \\
        Cross-attention \cite{hou2019cross} & 67.09 $\pm$ 0.55 & 80.64 $\pm$ 0.35 & - & - & - & - & - & - \\
        BD-CSPN \cite{liu2020prototype} & 81.89 $\pm$ 0.60 & 70.31 $\pm$ 0.93 & -& -& - & - &- \\ 
        PT+MAP \cite{hu2021leveraging} & - & - & 90.50 $\pm$ 0.49 & 86.91 $\pm$ 0.72 & 54.96$\pm$ 0.24 & 40.88$\pm$ 0.21 & 93.93 $\pm$ 0.32 & 91.37 $\pm$ 0.61 \\
        LR+ICI \cite{wang2020instance} &-&-& 89.75 $\pm$ 0.48 & 84.88 $\pm$ 0.79&-&-& 93.35  $\pm$ 0.30 & 90.18 $\pm$ 0.6\\ 
        DeepEMD \cite{zhang2020deepemd}&-&-&-&-&-&-& 88.69  $\pm$ 0.50 &  76.65  $\pm$0.83 \\
        TIM-GD \cite{boudiaf2020information} &-&-&-&-&-&-&  92.14  $\pm$ 0.10 & 88.35  $\pm$ 0.19 \\
        ODC \cite{qi2021transductive} & 87.11 $\pm$ 0.42 & 77.20 $\pm$ 0.36 & - & - & 57.63 $\pm$ 0.23 & 42.04 $\pm$ 0.17 & - & - \\
        EPNet \cite{qi2021transductive} & 84.34 $\pm$ 0.85 & 70.74 $\pm$ 0.85& -& -& - & - &- \\
        MDM-Net \cite{gao2022multi} &-&-&-&- & 57.41  $\pm$ 0.33 & 43.62  $\pm$ 0.46 &-
        &-\\
        Metaopt Net \cite{gong2023meta} &-&-&-&-&  55.50 $\pm$ 0.60 &  41.10 $\pm$ 0.60&-&-  \\ 
        iLPC \cite{lazarou2021iterative} & 88.82 $\pm$ 0.42 & 83.05 $\pm$ 0.79& 90.60 $\pm$ 0.48 & 86.51 $\pm$ 0.75 &-&-& 94.11 $\pm$ 0.30 & 91.03  $\pm$ 0.63 \\ 
        protoLP \cite{zhu2023transductive} & 90.45 $\pm$ 0.11  & 85.13 $\pm$ 0.24 & 91.52 $\pm$ 0.15 & 88.22 $\pm$ 0.21 & 57.50 $\pm$ 0.65 & 41.00 $\pm$ 0.21 & 94.65 $\pm$ 0.10 & 91.82 $\pm$ 0.18 \\
        SP-CLIP \cite{chen2023semantic} &83.04$\pm$0.30 &77.63$\pm$0.63&88.24$\pm$0.32&82.18$\pm$0.40&61.55$\pm$0.41&48.53$\pm$0.38&-&-\\
        SemFew\cite{zhang2024simple}&86.49$\pm$90.50&78.9$\pm$40.66&89.11$\pm$0.54&84.34$\pm$0.67&-&-&-&-\\\hline
        \textbf{LaHA (Ours)} & \textbf{94.92 $\pm$ 0.22} & \textbf{88.97 $\pm$ 0.31} &
        \textbf{94.98 $\pm$ 0.41} & \textbf{90.12 $\pm$ 0.27} & \textbf{64.82 $\pm$ 0.55} & \textbf{49.67 $\pm$ 0.49} & \textbf{96.21 $\pm$ 0.25} & \textbf{94.22 $\pm$ 0.24} \\ \hline
    \end{tabular}%
    }
\label{tab:results_all}
\end{table*} 
\section{Experiments}
We conducted experiments on four commonly employed datasets for few-shot learning, MiniImageNet \cite{ravi2016optimization},
CIFAR-FS \cite{krizhevsky2009learning}, FC-100\cite{krizhevsky2009learning}, and CUB \cite{welinder2010caltech}. For stable training, we first pre-train the CL module and baseline classifier on randomly selected patches before introducing the RL module. However, the ViT agent is trained from scratch, demonstrating the RL framework’s effectiveness without prior knowledge. This underscores the approach’s flexibility while allowing potential performance gains with a pre-trained ViT. Detailed descriptions of each dataset, hyperparameters, and implementation specifics are provided in the Appendix.

\subsection{Improving Few-shot Learning}\label{Improving Few-shot Learning}
We used Prototypical Networks (ProtoNet) \cite{snell2017prototypical}, Prototype-based Label Propagation (ProtoLP) \cite{zhu2023transductive}, HyperShot \cite{sendera2023hypershot}, and LaplacianShot \cite{ziko2020laplacian} as the baseline classifier models to assess the improvements LaHA introduces when applied.
Additionally, we chose MOLMO, an open-source family of vision-language models developed by the Allen Institute for AI\s{, that performed best in our experiments} \cite{deitke2024molmo}. MOLMO is trained on the PixMo dataset, which contains 1 million curated image-text pairs\s{, and achieves state-of-the-art performance among similar-sized multimodal models}. We also tested two other VLMs, BLIP \cite{li2023blip} and ViLT \cite{kim2021vilt}, to examine the impact of different VLMs on the performance; \b{the results are presented} in the Appendix.

The \b{few-shot classification} results\b{, based on mean and standard deviation over the few-shot learning \textit{episodes}}, are presented in Table \ref{tab:results}\b{, where LaHA\_X denotes the performance of the FSL baseline classifier X after its integration and training in our proposed LaHA framework. The backbones used in each of the baseline methods are shown in parentheses\s{ in front of their name}.} 
The first \b{four} rows of Table \ref{tab:results} show the test results of the \b{four} baselines X on the original data (without any selection), and the second \b{four} rows show the outcomes obtained when integrating X in the LaHA framework, i.e. few-shot classification is performed only using the attentive regions. As can be observed, LaHA consistently improves \b{the employed baseline's} performance. \b{Note that such improvement is obtained although LaHA reduces the image size to at least \b{$1/N_a$} of the image, besides providing interoperability. $N_a$ is set to 3 in this experiment. }
\s{\r{ REMOVE: This is particularly advantageous because, firstly, \b{LaHa selects only the important regions of the input image; this reduces the data size significantly. } selecting these patches reduces the data size significantly, and secondly, retaining essential regions enhances interpretability, indicating the effectiveness of the proposed method in identifying informative and discriminative patch locations.} }

To further evaluate \b{the few-shot classification performance of LaHA compared to the SoTA,} we adopt protoLP as LaHA's baseline classifier (where $N_a$ = 3) and compare the results with several state-of-the-art few-shot \b{classification} approaches on different datasets, \s{. Accuracy comparison on mini-ImageNet, CIFAR-FS, FC-100 and CUB datasets are presented} in Table \ref{tab:results_all}. As shown, LaHA outperforms all other methods by notable margins across all datasets, highlighting LaHA's effective use of hard attention to improve generalization and overall performance in few-shot learning.
\subsection{Ablation Study}
To assess the impact of the CL and VLM modules on our method’s performance, we conducted a series of experiments on our datasets using the four FSL baselines. We evaluated four conditions: (1) excluding both CL and VLM (i.e., $\xi_2 = 0, \beta = 0$), (2) excluding only the VLM module (i.e., $\xi_2 = 0$), (3) excluding only the CL module ($\beta = 0$), and (4) including both modules, i.e. the whole framework, together. The \b{classification} accuracy results \s{across all few-shot learning baselines} are presented in Table \ref{tab:abb}. The classification accuracy results show that while removing CL and VLM leads to the lowest performance, the model still outperforms not using LaHA at all (see Table \ref{tab:results}), highlighting that RL is the core driver of performance. Adding VLM or CL independently yields improvements, with the best results achieved when both are included. The CL module aids in learning better representations, which is particularly valuable in few-shot settings where limited samples make generalization more difficult. Meanwhile, incorporating the VLM module refines the RL reward function by injecting semantic information, further enhancing classification accuracy. However, these components act as auxiliary enhancements, while the primary performance gains come from RL. Further experimental analysis on VLM's visual impact is available in the Appendix.
\begin{table*}[ht!]
    \centering
        \caption{The accuracy results of the ablation study on the CL module and VLM.}
    \resizebox{\textwidth}{!}{%
    \begin{tabular}
    {c|c|l|c|c|c|c|c|c|c}
        \hline
        \multirow{4}{*}{} & \multirow{2}{*}{\textbf{Methods}} & \multicolumn{2}{c|}{\textbf{MiniImageNet}} & \multicolumn{2}{c|}{\textbf{CIFAR-FS}} & \multicolumn{2}{c|}{\textbf{FC-100}} & \multicolumn{2}{c}{\textbf{CUB}} \\ \cline{3-10}
         & & \textbf{5-s 5-w} & \textbf{1-s 5-w} & \textbf{5-s 5-w} & \textbf{1-s 5-w} & \textbf{5-s 5-w} & \textbf{1-s 5-w} & \textbf{5-s 5-w} & \textbf{1-s 5-w} \\ \hline
         & LaHA\_ProtoNet & 74.93 $\pm$ 0.25 & 56.64 $\pm$ 0.67 & 80.52 $\pm$ 0.67 & 69.01 $\pm$ 0.54 & 51.02 $\pm$ 0.81 & 35.33 $\pm$ 0.69 & 84.91 $\pm$ 0.42 & 71.23 $\pm$ 0.63 \\ 
        \b{No CL \& No VLM} & LaHA\_LaplacianShot & 84.13 $\pm$ 0.16 & 74.86 $\pm$ 0.19 & 86.42 $\pm$ 0.39 & 74.89 $\pm$ 0.31 & 55.87 $\pm$ 0.65 & 39.32 $\pm$ 0.69 & 88.68 $\pm$ 0.69 & 80.96 $\pm$ 0.69 \\ 
        ($\xi_2 = 0, \b{\beta} = 0
        $) & LaHA\_HyperShot & 69.62 $\pm$ 0.20 & 53.18 $\pm$ 0.50 & 78.20 $\pm$ 0.45 & 60.15 $\pm$ 0.55 & 50.30 $\pm$ 0.70 & 34.20 $\pm$ 0.60 & 80.07 $\pm$ 0.22 & 66.13 $\pm$ 0.26 \\ 
         & LaHA\_protoLP  & 92.22 $\pm$ 0.11 & 86.35 $\pm$ 0.24 & 92.52 $\pm$ 0.15 & 88.22 $\pm$ 0.21 & 57.50 $\pm$ 0.65 & 41.00 $\pm$ 0.70 & 94.65 $\pm$ 0.10 & 91.82 $\pm$ 0.18 \\ \hline
        & LaHA\_ProtoNet & 77.34 $\pm$ 0.32 & 59.21 $\pm$ 0.41 & 81.01 $\pm$ 0.58 & 71.02 $\pm$ 0.62 & 51.94 $\pm$ 0.72 & 38.24 $\pm$ 0.79 & 88.21 $\pm$ 0.41 & 73.82 $\pm$ 0.55 \\ 
        No VLM& LaHA\_LaplacianShot  & 86.23 $\pm$ 0.14 & 79.24 $\pm$ 0.23 & 85.34 $\pm$ 0.43 & 74.89 $\pm$ 0.31 & 57.00 $\pm$ 0.66 & 41.98 $\pm$ 0.69 & 92.42 $\pm$ 0.39 & 84.88 $\pm$ 0.45 \\ 
        ($\xi_2 = 0$)& LaHA\_HyperShot & 71.12 $\pm$ 0.23 & 55.12 $\pm$ 0.54 & 79.18 $\pm$ 0.45 & 62.72 $\pm$ 0.54 & 50.30 $\pm$ 0.70 & 36.02 $\pm$ 0.68 & 82.24 $\pm$ 0.32 & 67.21 $\pm$ 0.36 \\ 
        & LaHA\_protoLP  & 93.39 $\pm$ 0.11 & 87.32 $\pm$ 0.24 & 93.95 $\pm$ 0.35 & 89.63 $\pm$ 0.31 & 58.76 $\pm$ 0.61 & 44.12 $\pm$ 0.52 & 96.21 $\pm$ 0.25 & 91.82 $\pm$ 0.18 \\ \hline
        & LaHA\_ProtoNet & 76.34 $\pm$ 0.31 & 57.21 $\pm$ 0.41 & 80.93 $\pm$ 0.62 & 70.45$\pm$ 0.49 & 51.92 $\pm$ 0.79 & 38.14 $\pm$ 0.79 & 85.92 $\pm$ 0.49 & 71.44 $\pm$ 0.43 \\ 
        No CL& LaHA\_LaplacianShot & 87.36 $\pm$ 0.18 & 76.93 $\pm$ 0.29 & 85.97 $\pm$ 0.48 & 75.23 $\pm$ 0.37 & 56.92 $\pm$ 0.69 & 41.75 $\pm$ 0.64 & 90.56 $\pm$ 0.28 & 83.68 $\pm$ 0.49 \\ 
        ($\beta = 0$)& LaHA\_HyperShot & 70.38 $\pm$ 0.42 & 53.87 $\pm$ 0.48 & 78.78 $\pm$ 0.51 & 61.97 $\pm$ 0.52 & 51.12 $\pm$ 0.74 & 34.55 $\pm$ 0.68 & 81.22 $\pm$ 0.35 & 67.01 $\pm$ 0.33 \\ 
        & LaHA\_protoLP  & 92.95 $\pm$ 0.35 & 86.73 $\pm$ 0.43 & 93.55 $\pm$ 0.31 & 88.84 $\pm$ 0.31 & 58.76 $\pm$ 0.64 & 42.85 $\pm$ 0.61 & 95.43 $\pm$ 0.29 & 92.39 $\pm$ 0.31 \\ \hline
        & LaHA\_ProtoNet & \textbf{79.54 $\pm$ 0.13} & \textbf{61.29 $\pm$ 0.60} & \textbf{81.50 $\pm$ 0.65} & \textbf{71.02 $\pm$ 0.60} & \textbf{52.50 $\pm$ 0.80} & \textbf{39.00 $\pm$ 0.75} & \textbf{90.77 $\pm$ 0.41} & \textbf{75.59 $\pm$ 0.43} \\ 
        The original framework& LaHA\_LaplacianShot & \textbf{89.12 $\pm$ 0.24} & \textbf{82.02 $\pm$ 0.41} & \textbf{87.00 $\pm$ 0.44} & \textbf{75.80 $\pm$ 0.45} & \textbf{58.00 $\pm$ 0.70} & \textbf{43.00 $\pm$ 0.68} & \textbf{93.12 $\pm$ 0.25} & \textbf{86.02 $\pm$ 0.41} \\
        $\xi_2 \neq 0, \beta \neq 0$& LaHA\_HyperShot & \textbf{72.37 $\pm$ 0.26} & \textbf{55.98 $\pm$ 0.63} & \textbf{80.65 $\pm$ 0.60} & \textbf{63.87 $\pm$ 0.55} & \textbf{51.93 $\pm$ 0.65} & \textbf{37.98 $\pm$ 0.70} & \textbf{82.93 $\pm$ 0.31} & \textbf{68.14 $\pm$ 0.35} \\ 
        &LaHA\_protoLP  & \textbf{94.92 $\pm$ 0.22} & \textbf{88.97 $\pm$ 0.31} &
        \textbf{94.98 $\pm$ 0.41} & \textbf{90.12 $\pm$ 0.27} & \textbf{64.82 $\pm$ 0.55} & \textbf{49.67 $\pm$ 0.49} & \textbf{96.21 $\pm$ 0.25} & \textbf{94.22 $\pm$ 0.24} \\ \hline
    \end{tabular}%
    }

    \label{tab:abb}
\end{table*}

\subsection{Visual Evaluation}
\begin{figure}[t]
\centerline{\includegraphics[width=1.1\linewidth]{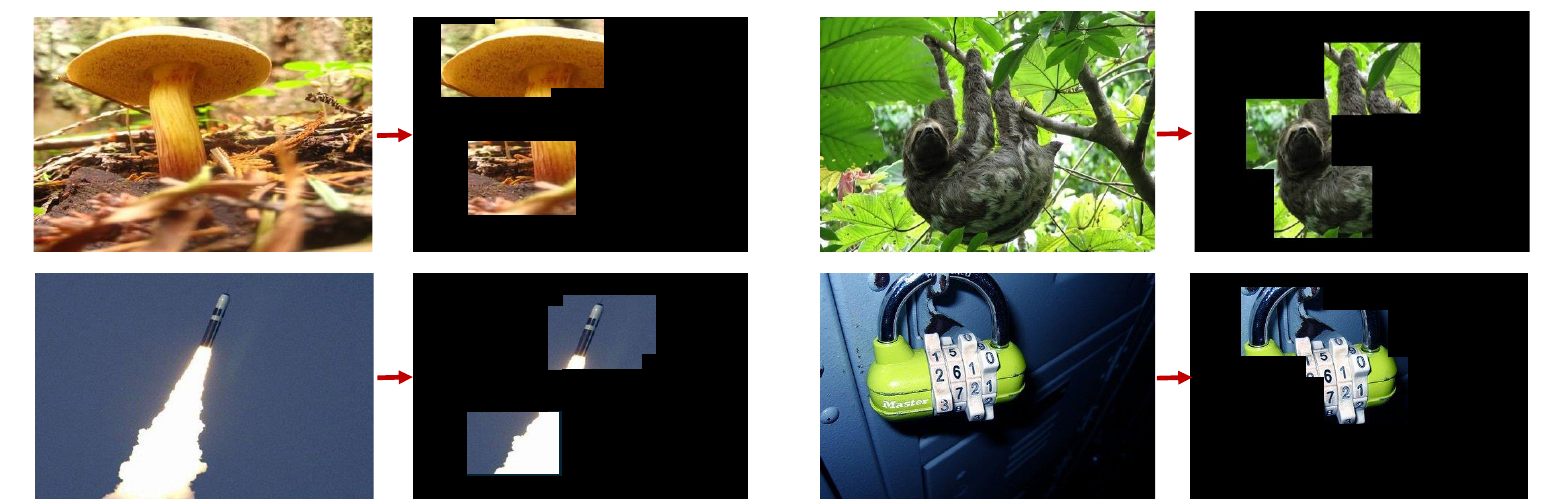}}
\caption{Visualization of the selected patches by LaHA.}
\label{fig:vis}
\end{figure}

\brown{LaHA's selected patches (with $N_a$ = 3), for a few samples of mini-ImageNet, using protoLP (ResNet-12) as the baseline, }are shown in Figure \ref{fig:vis}.  The visual output of LaHA affirms its consistency with human perception, contributing to the interpretability of the learning model. 
\b{Additionally, observing the overlapping patches in the images suggests that more information does not necessarily improve performance, and what matters is providing the model with the most relevant and non-redundant inputs.}
\s{Furthermore, the overlapping patches in the images indicate that increasing the patch size or number does not contribute additional useful information, meaning three is a reasonable number for the number of patches.}

\begin{figure}[t]
\centerline{\includegraphics[width=1.05\linewidth]{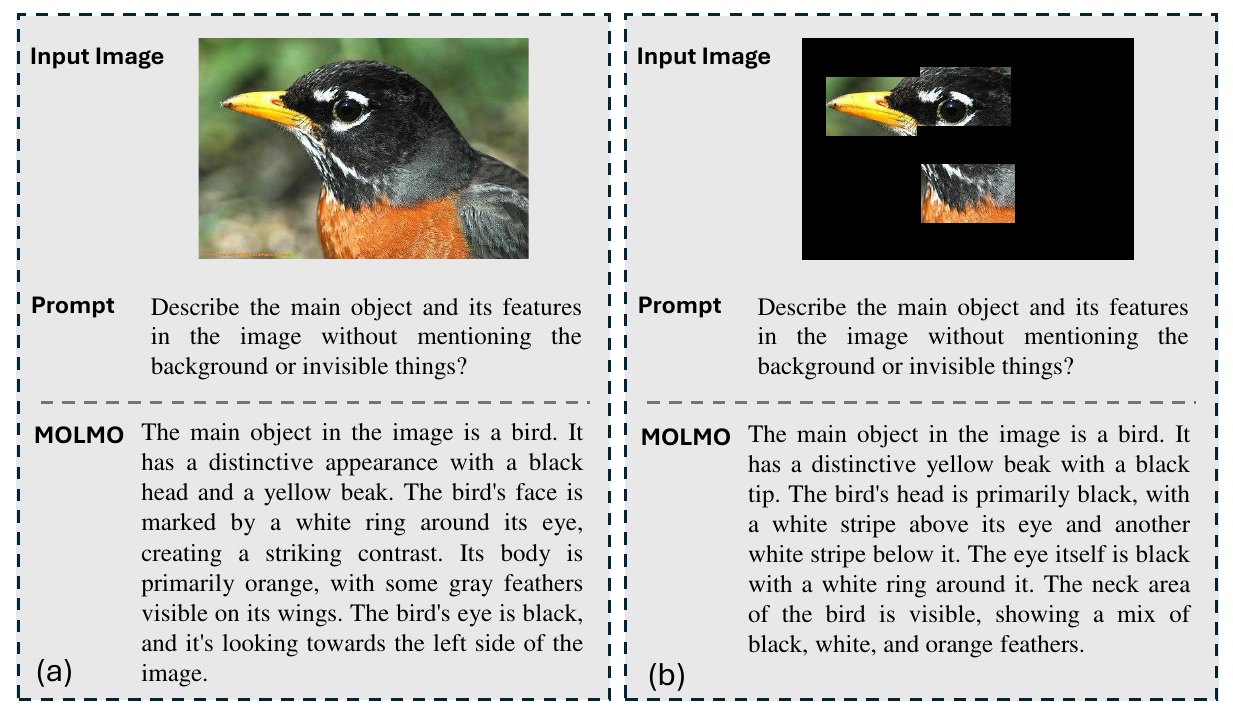}}
\caption{The output of VLM (MOLMO), (a) before and (b) after applying LaHA.}
\label{molmo:pd}
\end{figure}

\brown{To further evaluate whether the selected attentive patches retain important information, we compare the VLM's extracted content before and after selecting these patches, using a proper prompt given to MOLMO. Comparing the outputs shown in Figure \ref{molmo:pd} indicates that MOLMO consistently captures the same essential information, regardless of whether the full image or the attentive patches are provided, affirming LaHA's ability to focus on critical details without losing key semantic context. This consistency confirms that adding the VLM's reward leads to selecting regions aligned with human interpretation, making the LaHA's selection process more reliable.}

\subsection{Beyond few-shot learning}
To demonstrate the adaptability and effectiveness of our proposed LaHA framework beyond its original scope on few-shot learning, we extended our evaluation to the standard image classification task using the ILSVRC 2012 version of the ImageNet dataset, which comprises 1,281,167 training images and 50,000 validation images spanning 1,000 natural image classes \cite{russakovsky2015imagenet}. We selected ResNet-50 as the baseline classifier model for LaHA due to its proven performance in image classification tasks. LaHA was compared against three leading hard attention-based methods for classification: Saccader \cite{elsayed2019saccader}, DRAM \cite{ba2014multiple}, and Traversal Network (TNet) \cite{xu2018learning}, with detailed descriptions provided in the Appendix. The performance of each method under varying configurations of attentive regions (\b{$N_a = $} 2, 4, 5, and 6) is summarized in the table, with unavailable data indicated by “-”. Note that the performance results for $N_a$ = 3 are not reported for any of these methods in their respective papers. The results demonstrate that LaHA consistently outperforms other methods across all configurations, showing its robustness and precision in identifying critical image regions. \brown{Since the attentive regions in LaHA can overlap, even when $N_a$ is set to a larger value, the model still focuses on the most important regions. This means that increasing $N_a$ does not force the model to use all patches; instead, it learns to select only the most relevant ones. The results show that increasing $N_a$ beyond 4 does not improve performance, indicating that LaHA effectively removes redundant information while keeping classification accuracy high.
}
\begin{table}[t!]
\centering
\caption{Performance Comparison of Hard Attention Methods on ImageNet.}
\small
\begin{tabular}{l|c|c|c|c}
\hline
\textbf{Model} & \textbf{\b{$N_a=2$}} & \textbf{\b{$N_a=4$} } & \textbf{\b{$N_a=5$} } & \textbf{\b{$N_a=6$} } \\ \hline
Saccader \cite{elsayed2019saccader}  &  67.79\% & - & - & 70.31\%\\ \hline
DRAM \cite{ba2014multiple} &  49.72\% &  64.26\% &-&-\\ \hline
TNet \cite{xu2018learning} &  74.12\% &  74.41\%& 74.62 \%&- \\ \hline\hline
LaHA (Ours) & \textbf{77.32\%} & \textbf{78.88\%} &\textbf{79.06\%}&\textbf{79.02\%}\\ \hline
\end{tabular}
\end{table}

\section{Conclusion}

In this paper, we present LaHA, a novel approach designed to tackle the challenges of few-shot learning by leveraging deep reinforcement learning with Vision Transformers to identify hard attention regions directly from RGB images. Unlike soft attention methods, LaHA focuses on informative patches, reducing background noise and unnecessary data while enhancing interpretability. By incorporating a vision-language model as an additional reward signal, LaHA provides clearer insights into the model's decision-making process. Experiments conducted on multiple datasets confirm LaHA's efficacy in few-shot learning, showing both performance improvements and reduced data size, making it well-suited for resource-constrained environments. Also, LaHA exhibits adaptability to standard image classification tasks, underscoring its potential as a versatile tool for a broader range of computer vision applications where efficiency and interpretability are paramount.


{
    \small
    \bibliographystyle{ieeenat_fullname}
    \bibliography{main}
}

\end{document}